\documentclass[runningheads]{llncs}
\usepackage{graphicx}
\begin{document}
\title{FP-PET: Large Model, Multiple Loss And Focused Practice}
\author{Yixin Chen\inst{1}\orcidID{0000-0002-2727-6387} \and
Ourui Fu\inst{2} \and
Wenrui Shao\inst{1}\and
Zhaoheng Xie \inst{1,3}
}
\authorrunning{Y. Chen et al.}
\institute{ Institute of Medical Technology, Peking University, Beijing 100191, China. \\ \and 
Department of Biomedical Engineering, Peking University, Beijing 100191, China. \\ \and
National Biomedical Imaging Center, Peking University, Beijing 100191, China.
}

\maketitle              % typeset the header of the contribution
%
% \begin{abstract}
% The abstract should briefly summarize the contents of the paper in
% 15--250 words.

% \keywords{First keyword  \and Second keyword \and Another keyword.}
% \end{abstract}
%
%
%
\section{Data}
\subsection{Training/Validation Process}
In our study, we utilized the dataset provided by the AutoPet2023 Challenge, which comprises over 1,000 cases, approximately half of which are normal and the other half contain lesions \cite{autopetdata}. We don't pursue the 5-fold cross-validation technique for assessing the model due to the following two reasons:

\begin{enumerate}
    \item Computational Costs: The training overhead associated with 5-fold cross-validation would be significant, particularly as we planned to experiment with three distinct models: STUNet \cite{STUnet}, VNet \cite{vnet}, and SwinUNETR \cite{swinunetr}.
    \item Inference Time: Given that the inference for a single case could be time-consuming, performing 5-fold cross-validation would exacerbate the temporal constraints of our study.
\end{enumerate}

Thus, we opted for an alternative model evaluation approach. We randomly selected 30 lesion cases and 20 normal cases from the extensive dataset provided by the AutoPet2023 Challenge. Given the large size of the official dataset, the exclusion of 50 cases for validation is unlikely to significantly impact the training process. Furthermore, we generated three distinct validation sets, each consisting of 50 cases. Each validation set was designated for a specific model architecture.

This approach provides meaningful insights into the model's performance without incurring excessive computational costs or compromising the reliability of the evaluation results. We term this method as ``model-wise cross validation".

\subsection{Preprocess}
Due to hardware limitations, our memory resources were insufficient to load more than 1,000 cases of both CT and PET images simultaneously, particularly at a spacing resolution of $1.5mm\times 1.5mm \times 1.5mm$. To address this challenge, we cropped all data to $128\times128\times128$ patch size, using a 0.5 overlap between adjacent patches. Ultimately, this pre-processing step yielded a dataset of 100,000 patches for training.

In terms of image fusion between CT and PET, we simply concatenated the CT and PET image along the channel dimension, resulting in a tensor shape of $[batch,2,128,128,128]$ for the model input.

Additionally, we initially planned to use various data augmentation techniques such as affine transformations, elastic trasnformation, and noise addition. However, empirical testing  revealed unexpected $I\/O$ bottlenecks, possibly due to unspecified issues related to our CPU and server storage. These issues led to prohibitive time costs during the I/O processes, preventing effective coorindation with the GPU when utilizing the aforementioned augmentations. Consequently, we restricted our data augmentation strategy to random flip across the three dimensions. 

\subsection{Postprocess}

Our model is designed to process $128\times 128 \times 128$ patch size. Therefore, when performing inference on a complete sample, the sample is subdivided into multiple patches with a 0.5 overlap between adjacant patches. To merge these patches effectively, we employed a gaussian weighting scheme for the overlapping regions. This approach is predicted on the observation that the model's confidence in segmentation results often follows a gaussian distribution decline from center to edge. 

Additionally, we experimented with post-processing techniques, specifically erosion and dilation, to refine the segmentation output. Although these methods were effective in significantly reducing the false-positive volume (FPV), they simultaneously increased the false-negative volume (FNV). As a result, we decide to exclude erosion and dilation from our final segmentation process..

\section{Focused Practice}
We introduce a strategy called Focused Practice (FP) aimed at accelerating model convergence and improving training performance. Specifically, we monitor the loss for each training patch and categorize these patches into ``hard" and ``easy" samples. A hard patch is defined as a sample that poses a particular challenge for the model, while an easy patch is one that the model can readily handle. This concept of hard-easy sampling has previously been employed in neural network training, such as focal loss \cite{focalloss}. We opted not to use focal loss in order to avoid introducing additional weighting factors into our loss function. Our current loss function already incorporates multiple loss terms, and we found that focal loss was not necessary for achieving our desired results. This could introduce unpredictability and potential instability during training. For instance, over-weighting an exceptionally challenging sample could disrupt the model's parameters.

Our FP strategy involves dynamically oversampling hard patches. The term ``dynamic" implies that the range of hard samples evolves in conjunction with ongoing model updates. FP continually adapts to model changes, which makes it an effective mechanism for oversampling hard samples. We determine the threshold for classifying hard and easy samples based on maximum inter-class variance. 

Furthermore, we avoid incorporating extremely hard samples into our training process for several reasons:

\begin{enumerate}
    \item During the patch cropping process, extremely small lesion areas might be included in a single patch. Overlooking these minute positive areas could disproportionately inflate the Dice loss.
    \item The learning trajectory of the model naturally progresses from easy to hard samples. Wtih the model gradually proficient with hard samples, extremely hard samples are downgraded to simply hard.
    \item Over-focusing on extremely challenging samples could cause the model to forget easy samples, leading to a situation where easy samples become hard in the next epoch, thereby destabilizing the training process.
\end{enumerate}

Consequently, we exclude the 20\% of hard samples with the highest loss and incorporate the remaining 80\% into the training data for the subsequent epoch.

\section{Experiments}
\subsection{Local Evaluation}
We employed the official metrics code provided to evaluate our model's performance on the validation set using three key metrics: Dice coefficient, false positive volume (FPV), and false negative volume (FNV). To synthesize these metrics into a comprehensive measure, we designed an aggregated score. The formula for calculating this score is as follow:

\begin{equation}
    score = dice-0.1*FPV-0.1*FNV
\end{equation}

\subsection{Implementation Details}
The model training was conducted across two servers. One node was equipped with four NVIDIA A800 GPUs, while the other node housed six NVIDIA 4090 GPUs.

\subsection{Model Structure}
For the architectural framework of our model, we opted for STUNet-large \cite{STUnet}, SwinUNETR\cite{swinunetr}, and VNet\cite{vnet}. The STUNet-large is a computationally intensive 3D segmentation model that has demonstrated competitive state-of-the-art performance in our ongoing research. Therefore, we plan to employ STUNet-large as the primary model, while incorporating SwinUNETR and VNet into the final ensemble with smaller weighting factors.

SwinUNETR, featuring a transformer architecture, diverges significantly from traditional Convolutional Neural Networks (CNNs). We believe that this divergence will contribute to enhanced generalizability. Meanwhile, VNet serves as a derivative of the UNet framework, representing traditional medical imaging segmentation models within our ensemble approach.

\subsection{Optimizer}
We experimented with three different optimizers: Stochastic Gradient Descent (SGD), Adam, and AdamW. For SGD, we initialized the learning rate at $3e-5$ with a momentum of 0.99 and a weight decay of $3e-5$. Both Adam and AdamW were configured with the same initial settings.

Through our experiments, we observed that SwinUNETR converged more quickly when optimized using Adam and AdamW. In contrast, SGD produced superior performance for STUNet and VNet.

We also implemented a straightforward dynamic learning rate strategy, as described below:

\begin{equation}
    lr = 3e-5 * 0.9 ^ {(dice/10)}
\end{equation}

where ``dice" is the dice coefficient of validation set of current epoch. We employed a dynamic learning rate strategy that adjusts in accordance with model accuracy improvements. Specifically, the learning rate decreases as the model exhibits improved performance, assessed by measuring the Dice coefficient on the validation set. Importantly, this approach is metric-driven and remains uninfluenced by variations in iteration counts per epoch and convergence rates, which may speed up because of different models or batch sizes.

To illustrate the advantages of our metric-driven dynamic learning rate strategy, we provide the following example. The convergence speed of VNet exceeds that of STUNet-Large, which in turn is faster than SwinUNETR. Consequently, SwinUNETR requires a greater number of epochs for training. Premature decay of the learning rate would result in an exceedingly lengthy training period for SwinUNETR. Additionally, due to the lower computational complexity of the VNet model, it allows for a larger batch size $32$, as compared to 6 for STUNet-Large, when using the same graphics card. Therefore, for a given epoch, VNet will undergo fewer iterations. Our metric-driven dynamic learning rate decay strategy effectively accommodates these variations, making it well-suited for application across diverse models.
\subsection{Loss Function}
We experimented with a variety of loss functions, including Hausdorff Loss \cite{HausdorffDTLoss}, Asym. Loss \cite{asymloss}, Tversky Loss \cite{TverskyLoss}, Soft Dice Loss \cite{vnet}, Dice Loss, and Cross-Entropy. Through our evaluations, we found that Soft Dice Loss outperformed Dice Loss. Asym. Loss did not yield noticeably better results, while Tversky Loss effectively reduced both the false positive volume (FPV) and false negative volume (FNV). Although we anticipated that Hausdorff Loss would be highly suitable for this task, we encountered unspecified issues that prevented its smooth deployment during the training phase. Consequently, we settled on a combination of Cross-Entropy, Soft Dice Loss, and Tversky Loss as our final choice for the loss functions.
\section{Limitations}
While our algorithm presents a notable advancement in PET/CT image segmentation, there are several limitations that we acknowledge and aim to address in future research:

\begin{enumerate}
    \item Tumor Type Identification: Our current algorithm does not distinguish between different types of tumors. Future version could aim to categorize tumors based on their histological or morphological characteristics.
    \item Image Augmentation: The training process would benefit from incorporating a more diverse set of image augmentation techniques, especially the addition of various types of noise, to enhance the model's generalizability.
    \item Incorporation of Boundary Loss: Although untested in our current model, we hypothesize that the use of boundary loss could significantly reduce both false positive volume (FPV) and false negative volume (FNV).
    \item Joint CT and PET Processing: Our algorithm lacks a module designed to jointly process CT and PET images, which could potentially improve the robustness and accuracy of lesion segmentation.
\end{enumerate}
\bibliographystyle{splncs04}
\bibliography{mybibliography}
\end{document}